\documentclass[runningheads]{llncs}

\usepackage{eccv}

\usepackage{eccvabbrv}

\usepackage{booktabs}
\usepackage{listings}
\usepackage{algorithm}
\usepackage{algorithmic}
\usepackage{tabularx}
\usepackage{lipsum} 
\usepackage{multicol}

\usepackage{multirow}
\usepackage{graphicx}
\PassOptionsToPackage{table,xcdraw}{xcolor}
\usepackage{xcolor}
\usepackage{colortbl}
\usepackage{multicol}

\usepackage[accsupp]{axessibility}  

\usepackage{hyperref}
\usepackage{orcidlink}

\begin{document}

\title{GRIDS: Grouped Multiple-Degradation Restoration with Image Degradation Similarity} 

\titlerunning{GRIDS: Grouped Multiple-Degradation Restoration}

\author{Shuo Cao\inst{1,2}\thanks{Equal contribution.\quad $\dagger$ Corresponding author.} \and
Yihao Liu\inst{2}\textsuperscript{\small\textasteriskcentered} \and
Wenlong Zhang\inst{2} \and
Yu Qiao\inst{2,3} \and
Chao Dong\inst{2,3,4}\textsuperscript{$\dagger$}  }

\authorrunning{S.~Cao et al.}

\institute{University of Science and Technology of China \and Shanghai Artificial Intelligence Laboratory \and
Shenzhen Institute of Advanced Technology, Chinese Academy of Sciences \and
Shenzhen University of Advanced Technology\\
\email{caoshuo@pjlab.org.cn}, \email{liuyihao14@mails.ucas.ac.cn},  \email{chao.dong@siat.ac.cn}}

\maketitle

\begin{abstract}
  Traditional single-task image restoration methods excel in handling specific degradation types but struggle with multiple degradations. To address this limitation, we propose Grouped Restoration with Image Degradation Similarity (GRIDS), a novel approach that harmonizes the competing objectives inherent in multiple-degradation restoration. We first introduce a quantitative method for assessing relationships between image degradations using statistical modeling of deep degradation representations. This analysis facilitates the strategic grouping of similar tasks, enhancing both the efficiency and effectiveness of the restoration process. Based on the degradation similarity, GRIDS divides restoration tasks into one of the optimal groups, where tasks within the same group are highly correlated. For instance, GRIDS effectively groups 11 degradation types into 4 cohesive groups. Trained models within each group show significant improvements, with an average improvement of 0.09dB over single-task upper bound models and 2.24dB over the mix-training baseline model. GRIDS incorporates an adaptive model selection mechanism for inference, automatically selecting the appropriate grouped-training model based on the input degradation. This mechanism is particularly useful for real-world scenarios with unknown degradations as it does not rely on explicit degradation classification modules. Furthermore, our method can predict model generalization ability without the need for network inference, providing valuable insights for practitioners.
\end{abstract}

\section{Introduction}
\label{sec:intro}
Images often encounter various forms of degradations during acquisition, transmission, or storage phases, deteriorating their visual quality and fidelity. Such degradations may appear as noise, blur, compression or artifacts. Consequently, there is a burgeoning demand for advanced image restoration techniques. Over the past decades, significant progress has been made in single-type restoration tasks, such as image super-resolution~\cite{srcnn,ranksrgan,chen2023activating}, denoising~\cite{dncnn,FFDNet}, deblurring~\cite{SRN,dv2,li2022,liu2023degae}, inpainting~\cite{wang2018image,li2022mat,wang2023self}, etc. However, these task-specific methods are limited to certain degradation types, which restricts their applicability and requires user expertise. This limitation emphasizes the need for more versatile and user-friendly approaches for multiple-degradation restoration.

\begin{figure}[t]
  \centering
   \includegraphics[width=\linewidth]{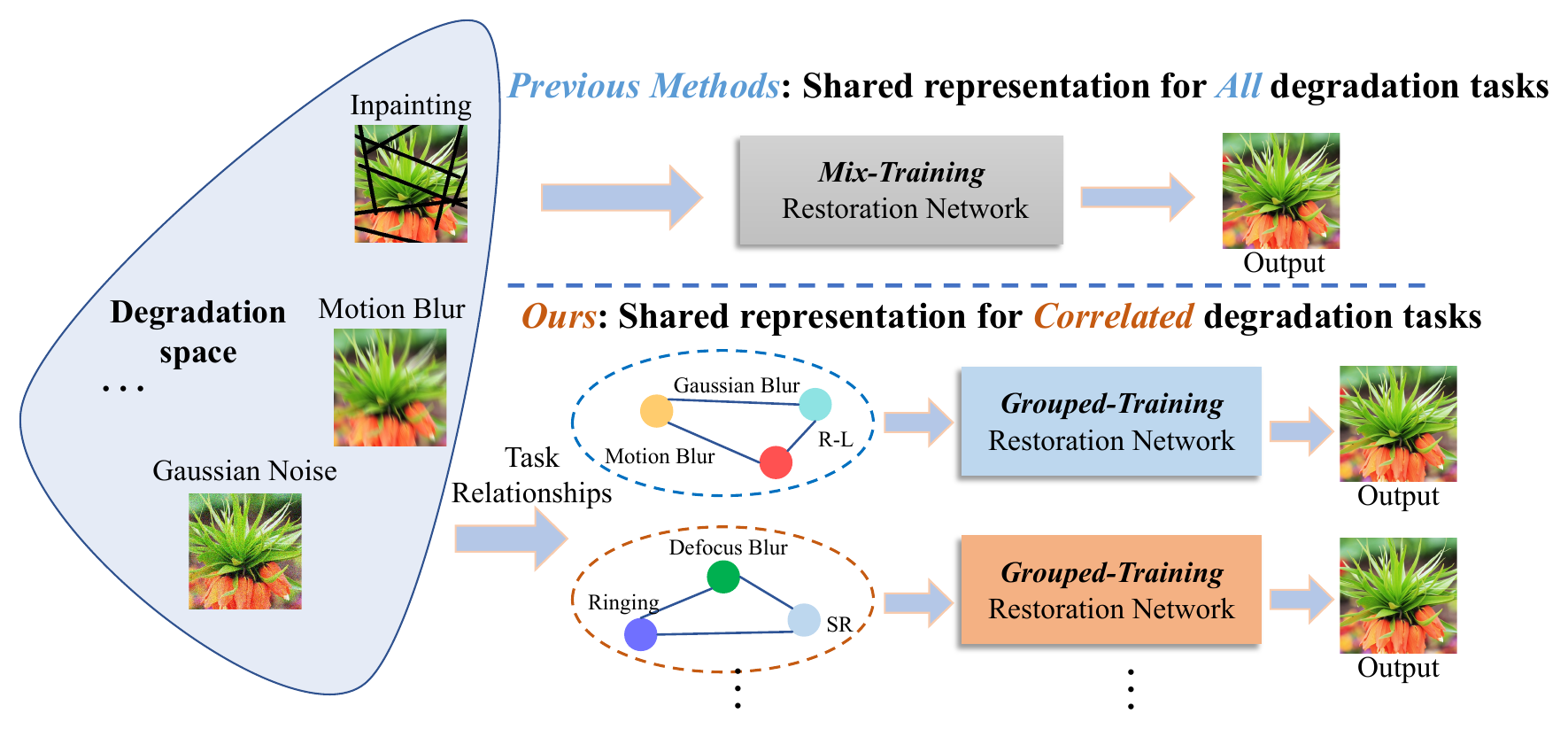}
   \caption{Schematic comparison of our GRIDS with the previous all-in-one methods.}
   \label{fig:intro}
\end{figure}

Multiple-degradation image restoration can be regarded as a multi-task learning (MTL) problem, where each degradation is treated as a separate task. By training a single model to handle all degradations, the model is supposed to demonstrate better generalization ability. However, as shown in Fig.~\ref{fig:intro}, this naive approach~\cite{zhang2022closer,AirNet,zhang2023real,liu2023unifying} is difficult to build a shared representation that can generalize to all tasks, inevitably leading to compromised performance for individual tasks (see Fig.~\ref{fig:motivation} and Tab.~\ref{tab:results_div2k}). Thus, it is crucial to analyze the relationships between different tasks in MTL. If tasks are closely related, the features learned for one task can be beneficial for the others. When tasks are negatively correlated or have conflicting objectives, optimizing them simultaneously will lead to a drastic performance drop for all tasks. From an MTL perspective, identifying task groups that benefit from the positives of joint-training while mitigating the negatives often improves the performance of multi-task learning systems~\cite{ruder2017overview,zhang2021survey,fifty2021efficiently}.

Existing methods for multiple-degradation restoration often overlook the relationships between tasks, resulting in unsatisfactory performance or limited generalization. One group introduces diverse and complex degradation types (e.g., blur, noise, resize, and JPEG) to create a sophisticated degradation model, such as BSRGAN \cite{bsrgan} and RealESRGAN \cite{realesrgan}. These methods strive to cover a wider range of real-world scenarios by combining various degradation types. However, their reliance on a shared representation for all tasks often leads to a performance drop on single degradation tasks~\cite{zhang2022closer,zhang2023real}. Another group \cite{dasr,AirNet,zhang2023ingredient,zhang2023all} discerns different degradation types and then apply degradation-specific processing in an all-in-one manner. To achieve this, they learn degradation-related embeddings to dynamically guide restoration. However, the all-in-one methods only consider a limited number of tasks (e.g., three or four tasks) and lack statistical quantitative description, resulting in inaccurate and limited degradation representations.

In fact, we observe that there is a trade-off between model generalization and performance. If we force a single model to deal with more kinds of degradations by simply mixing training data, the model will inevitably lead to declined performance on individual tasks. This is because different degradation types have their intrinsic characteristics, and training them together may lead to conflicts. On the contrary, if we seek for better performance on individual tasks, training a specialized single-task model is the best choice. Nonetheless, this requires training and storing a large number of models, and users have to keep switching between models based on different inputs. Apparently, there comes a dilemma among model generalization, task performance and number of models.

In this paper, we aim to mitigate such a dilemma from a multi-task learning perspective. Concretely, we present a quantitative method to measure the relationships between image degradations through the statistical modeling of deep degradation representations \cite{ddr}. Based on such degradation (task) relationships, we devise an approach for \textbf{G}rouped \textbf{R}estoration with \textbf{I}mage \textbf{D}egradation \textbf{S}imilarity (\textbf{GRIDS}), which divides tasks into one of the optimal groups. Tasks within the same group are highly correlated, thus can be trained using a single model without a significant performance decline. In our experiments, we take 11 degradation types as examples and divide them into 4 groups, achieving a compression rate of 63.6\%. The trained models within each group even yield a 0.09dB improvement on average compared to specialized models (upper bound), while surpassing the baseline model (mixing all data) by 2.24dB on average. At the inference stage, we design an adaptive model selection mechanism that can automatically select the appropriate group for testing according to the input degradation. 
More interestingly, our method can predict the model generalization ability and processing performance even without the need for network inference. This capability aids decision-making in real-world scenarios, particularly when dealing with unknown degradations.





Our contributions can be summarized as follows: (1) For the first time, we quantitatively measure the relationships between image degradations, providing new insights and tools for image restoration research. (2) Based on the degradation similarity matrix, we propose a task grouping-based approach called GRIDS for multiple-degradation restoration. GRIDS exceeds the baseline model by up to 5.23dB and on average by 2.24dB. (3) For inference, we devise an adaptive model selection mechanism that can automatically select the suitable group for testing. More interestingly, our method can predict the model's processing performance on arbitrary input images without actual inference. 


\section{Related Work}
\label{sec:related_work}


\noindent\textbf{Single-degradation restoration} has achieved significant progress in recent years, such as denoising~\cite{FFDNet,Gu_2019_ICCV,Uformer,chen2023masked}, deblurring~\cite{zhang2020deblurring,Restormer,li2022learning}, super-resolution~\cite{rcan,swinir,hat,dat}, etc. These approaches formulate new methodologies specifically for task-oriented datasets, achieving high performance on the same-domain test data. However, their generalization ability remains challenging.
\noindent\textbf{Multiple-degradation restoration} involves training a single model for multiple restoration tasks. There are two main categories: data-centric, represented by BSRGAN~\cite{bsrgan} and RealESRGAN~\cite{realesrgan}, and model-centric, represented by DASR~\cite{dasr} and AirNet~\cite{AirNet}. The former employs a single network to learn a wider range of tasks from a complex degradation model. The latter captures diverse degradations for specific processing~\cite{zhang2023ingredient,zhang2023all}. ProRes~\cite{ProRes}, PromptIR~\cite{PromptIR} and PromptGIP~\cite{liu2023unifying} extend this by using prompts for guidance, enabling MTL for more tasks. Nevertheless, these existing methods are still limited in performance or generalization. In contrast, our proposed approach breaks this ceiling by MTL across more than ten distinct tasks, while yielding competitive performance compared to single-task models.


\noindent\textbf{Multi-task learning} methods are categorized into task balancing, task grouping, and architecture design. Task balancing methods in ~\cite{guo2018dynamic, PCGrad, CAGrad, MGDA, guangyuanrecon, Graddrop} address task conflicts by adjusting the loss weights or modifying the update gradient. Task grouping ~\cite{Taskonomy,Wich_tasks_should_be_learned, shen2021variational, fifty2021efficiently} methods aim to determine which tasks should be learned together. For instance, Zamir et al.~\cite{Taskonomy} proposed a task taxonomy capturing task transferability, while TAG~\cite{fifty2021efficiently} computed affinity scores to group tasks. Architecture design methods are divided into hard and soft parameter-sharing methods. Hard parameter-sharing methods~\cite{kokkinos2017ubernet, long2017learning, bragman2019stochastic} use different decoders for each task, while soft parameter-sharing methods~\cite{misra2016cross,ruder2019latent,Nddr-cnn} allow cross-talk between task networks. Our method, inspired by task grouping and hard parameter sharing, uses degradation similarity to group degradation tasks and selects a specific network for each group to improve the overall performance.

\section{Motivation}\label{sec:motivation}

\begin{figure}[htbp]
    \centering 
    \begin{minipage}[t]{0.59\textwidth}
      \centering
       \includegraphics[width=\textwidth]{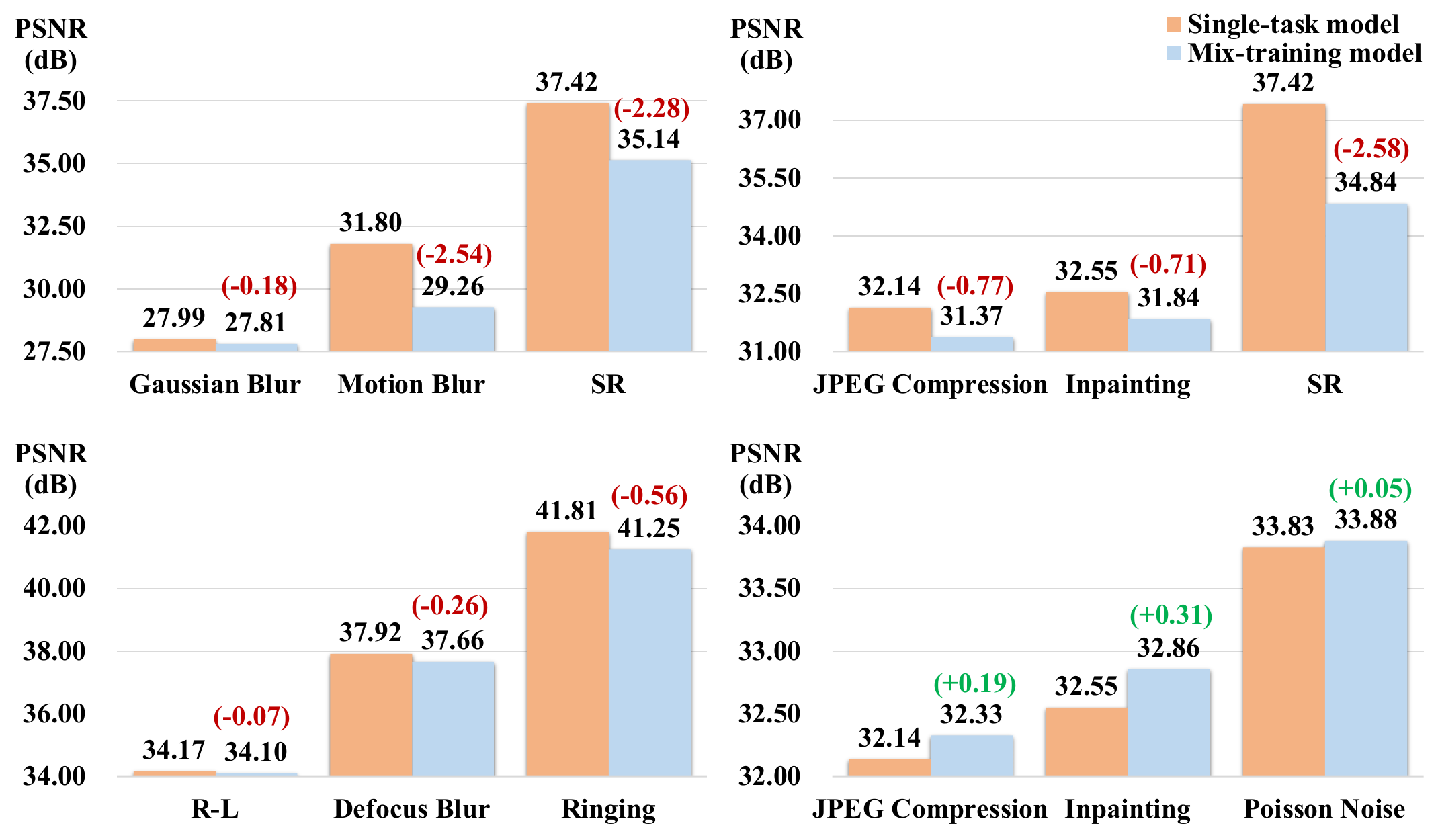}
       \caption{When tasks are negatively correlated, optimizing them simultaneously often results in a significant performance decline across all tasks. Conversely, training closely related tasks together does not lead to severe performance deterioration.}
       \label{fig:motivation}
    \end{minipage}
    \hfill 
    \begin{minipage}[t]{0.40\textwidth}
      \centering
       \includegraphics[width=\textwidth]{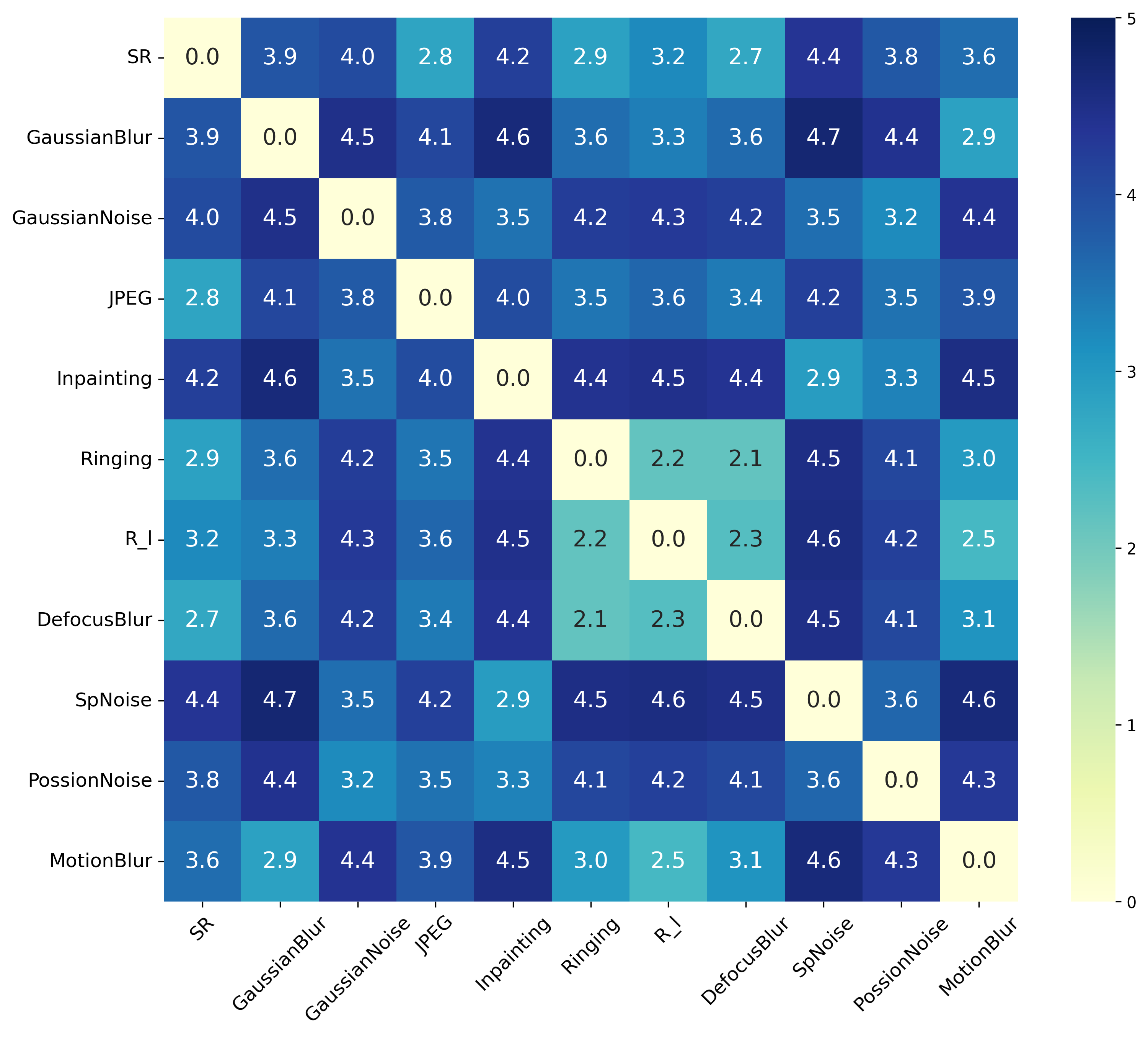}
       \caption{The obtained degradation similarity matrix $S$, which quantitatively describes the similarity relationship between 11 different degradations.}
       \label{fig:similarity_matrx}
    \end{minipage}
\end{figure}

\noindent\textbf{Performance and Generalization Trade-off.} Earlier methods attempt to train a single model for all tasks by mixing degradations during training. Unfortunately, simply increasing the number of degradations would result in a significant decline in performance. This decline is due to potential conflicts between unrelated tasks that are jointly optimized. For instance, when considering degradations like Gaussian blur, motion blur and SR, training a mix-training model exhibits severe performance drops (up to 2.54 dB) compared to specialized models, as shown in Fig.~\ref{fig:motivation}. This decline becomes more pronounced with an increased number of degradations. A baseline model trained on 11 different degradations shows drastic performance drops up to 5.32dB (see Tab.~\ref{tab:results_div2k}). Although training with various degradations can enhance generalization, it significantly compromises individual task performance. There is a noticeable trade-off between model performance and generalization.

\noindent\textbf{Relationships Between Image Degradations.} Understanding the interrelationships between tasks is crucial in multi-task learning. Different degradations exhibit distinct patterns and require varying features for restoration. Combining unrelated degradations together can lead to interference and competition, preventing the model from accurately learning task-specific features. Consequently, this compromised learning negatively impacts the performance of all tasks, resulting in the performance-generalization trade-off. However, previous methods have rarely explored the relationships between degradations. Our analysis reveals that by strategically grouping similar degradations together, such as defocus blur, ringing, and R-L degradations, the performance deterioration can be marginal (maximum 0.57dB, average 0.30dB). In some cases, like combining inpainting, Poisson noise, and JPEG compression, the performance for each task can even slightly improve (see Fig.~\ref{fig:motivation}).

These insights have motivated us to investigate the relationships between various degradations and develop a new approach for multiple-degradation restoration. In Sec.~\ref{sec:similarity}, we aim to identify the relationships between various degradations, providing references for subsequent model design. In Sec.~\ref{sec:GRIDS}, we devise a task grouping algorithm to divide various degradations into correlated groups, which can minimize interference and optimize the learning process.



\section{Identifying Degradation Relationships}\label{sec:similarity}
\subsection{Revisiting Deep Degradation Representations}
In a recent study, deep degradation representations (DDR)~\cite{ddr} are discovered within super-resolution (SR) networks. It is observed that a well-trained SR network (trained on the classic bicubic downsampling setting) inherently acts as an effective descriptor of image degradations. These deep features possess discriminative properties towards various degradation types, opening new possibilities for degradation analysis. Building upon this, another work~\cite{srga} introduces a generalization assessment measure for blind SR networks, namely SRGA. Specifically, SRGA exploits a generalized Gaussian distribution to model the deep representations of blind SR networks, where the distribution deviations of different inputs are quantified to indicate the generalization ability. Inspired by these pioneering works, we further explore the deep relationships between image degradations.

\subsection{Quantifying Image Degradation Similarity}
Measuring the relationships between image degradations can be challenging due to their diverse nature and context-dependent interactions. Fortunately, the presence of deep degradation representations offers a valuable opportunity to establish a statistical framework for quantifying the similarity between different degradations. In this study, we consider a set of $n$ degradations denoted as $\{d_i\}_{i=1}^{n}$. We first apply each degradation on PIES-Clean dataset~\cite{srga} to obtain the corresponding degraded image sets. Referring to~\cite{ddr}, we then utilize a pretrained SRResNet~\cite{srgan} as a degradation descriptor to extract the representations for each degradation. Following the methodology described in~\cite{srga}, we employ the generalized Gaussian distribution (GGD) to fit the degradation representations, estimating the distribution parameters $\alpha$ and $\sigma$:
\begin{equation}
GGD(x;\alpha, \sigma^2) = \frac{\alpha}{2 \beta \Gamma(1/\alpha)}\rm{exp}\left(-\left(\frac{|x|}{\beta}\right)^\alpha\right),
\end{equation}
where $\beta=\sigma \sqrt{\frac{\Gamma(1/\alpha)}{\Gamma(3/\alpha)}}$
and $\Gamma(z)=\int_{0}^{\infty} t^{z-1}e^{-t}\ dt \ (z>0)$.

Once the distribution for each degradation is obtained, we compute the logarithm of the Kullback-Leibler (KL) divergence as the similarity measure. The resulting similarity/distance matrix $S$ provides insights into the relationships between different degradations. The element $S_{ij}$ denotes the distance between the $i$-th and $j$-th degradations. Smaller $S_{ij}$ suggests these two degradations are more correlated. For example, as shown in Fig.~\ref{fig:similarity_matrx}, it is observed that ringing, R-L, defocus blur and motion blur degradations are relatively close to each other, while the Gaussian blur, Salt\&pepper noise and JPEG compression degradations exhibit significant variation. Remarkably, our method is the first to enable quantitative measurement of the similarity between degradations. 


It should be noted that due to the complexity of image degradation, coupled with the varying influences of image content, the obtained similarity matrix may exhibit certain uncertainties. In other words, the similarity scores between degradations may not strictly follow a monotonic pattern locally. Despite this inherent uncertainty, we have observed that the overall similarity matrix aligns well with experimental results. These similarity scores provide a valuable reference for understanding the degradation relationships, laying the groundwork for advancing more effective approaches in multiple-degradation restoration.




\section{Similarity-Based Task Grouping for Multiple-Degradation Restoration}\label{sec:GRIDS}


As presented in Sec.~\ref{sec:similarity}, the degradation similarity matrix $S$ can provide a valuable reference for categorizing different degradations into similar groups. While it is possible to manually group degradations using the similarity matrix $S$, this process is cumbersome and error-prone, especially when dealing with numerous degradation types with complex relationships. 

To address this challenge, we take a step further and introduce an approach for Grouped Restoration with Image Degradation Similarity (GRIDS). We formulate the multiple-degradation restoration task as a grouping optimization problem and present a heuristic binary search algorithm to autonomously identify the plausible solutions. Benefiting from the preprocessing based on similarity matrix, the search algorithm only needs to operate limited training and testing.

Furthermore, GRIDS incorporates an adaptive model selection mechanism that intelligently determines the most appropriate group for a given image. More intriguingly, thanks to the statistical modeling of degradations, our method can predict the model's performance on arbitrary inputs without actual inference.


\begin{figure*}[t]
  \centering
   \includegraphics[width=\linewidth]{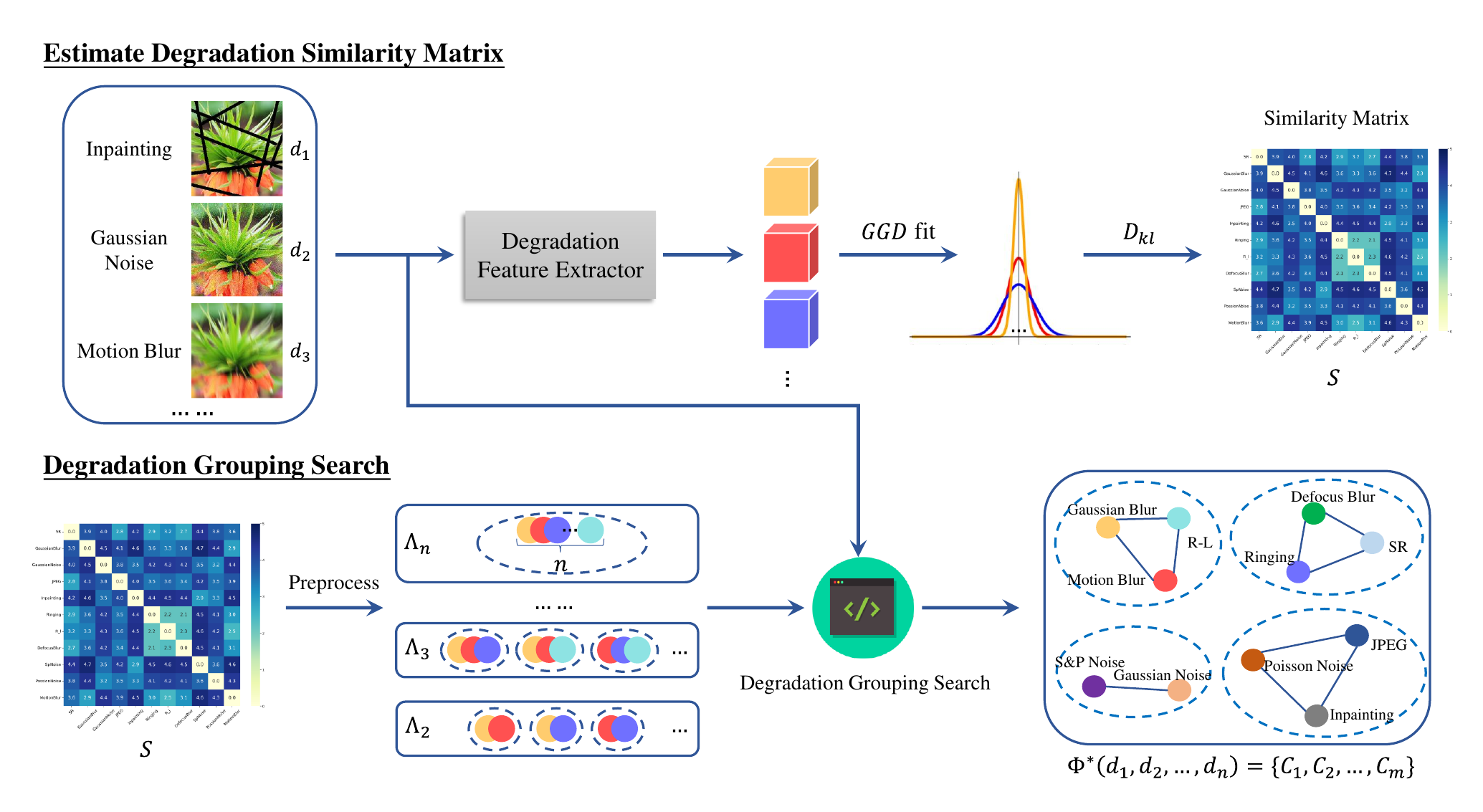}
   \caption{Workflow of the similarity-based degradation grouping method. Initially, a pretrained degradation feature extractor is employed to extract deep degradation representations, which are then modeled using the Generalized Gaussian Distribution. By computing the KL divergence between these representations, a similarity matrix is generated. Subsequently, the degradation grouping search algorithm utilizes the similarity matrix to automatically classify the degradations into relevant groups.}
   \label{fig:overview}
\end{figure*}

\subsection{Task Grouping with Degradation Similarity}
\noindent\textbf{Problem Formulation.}
Given $n$ degradation types $\{d_1, d_2, \ldots, d_n\}$, our objective is to find a partitioning scheme $\Phi$ to divide them into $m$ groups:
\begin{equation}
\Phi(d_1, d_2, \ldots, d_n)=\{C_1, C_2, \ldots, C_m\},
\end{equation}
where the $k$-th group contains $n_k$ degradation types:

\begin{equation}
C_k=\{d_{k_i}\}_{i=1}^{n_k}, k\in \{1,2,3,...,m\}.
\end{equation}

Intuitively, we hope that $\Phi$ has the minimum number of groups $m$, while maintaining the performance of each group at a high level. To formulate this objective, we first define some quantitative measurements.

For each degradation $d_i$, we can train a single-task model $M_S(d_i)$ as the upper bound model, since it is dedicated to specific degradation. Assuming we have obtained a degradation group $C_k$, we can train a mix-training model $M_M(C_k)$ for all degradations in $C_k$ using the same network architecture. For a trained model $M$ (single-task model or mix-training model), we define the performance of the model $M$ on degradation $d_i$ as $P(M, d_i)$. For example, $P(M(\{d_1,d_2\}),d_1)$ denotes the model performance on $d_1$, with training on $d_1$ and $d_2$. Then, we can define the maximum performance drop of group $C_k$ as:
\begin{equation}
\begin{aligned}
\Delta P(C_k) &= \max_{d_i\in C_k}{(P(M_S(d_i), d_i) - P(M_M(C_k), d_i))}.
\end{aligned}
\end{equation}


$\Delta P(C_k)$ reflects the maximum performance drop between mix-training model $M_M(C_k)$ and single-task models $M_S(d_i)$ across all degradations $d_i$ in group $C_k$.

Formally, we abstract the original problem into an optimization problem as

\begin{equation}
\begin{aligned}
&\min_\Phi m = |\Phi(d_1, d_2, \ldots, d_n)|, \\
s.t. &\Delta P(C_k) \leq \delta, k\in \{1,2,3,...,m\},
\end{aligned}
\label{eq:problem description}
\end{equation}
where $\delta$ is a manually-set threshold and $| \cdot |$ denotes the set cardinality. The optimization object suggests finding a partitioning scheme $\Phi$ that can divide all degradation types into as few groups as possible, and ensure that the performance drop is within the threshold $\delta$.






\noindent\textbf{Similarity Preprocessing.} 
Thanks to degradation similarity matrix $S$, we can group different degradations in a more systematic manner, rather than simply relying on intuition and observation. We devise an algorithm to efficiently search for all plausible solutions to Eq.~\ref{eq:problem description}, while minimizing training and testing costs.

Each potential group can contain $1$ to $n$ degradations. We denote the set of the groups containing $L$ degradations as $\Lambda_L$. For each group in $\Lambda_L$, we evaluate its average similarity and variance based on the similarity matrix $S$. Groups are first sorted by average similarity from highest to lowest (distance smallest to largest). For nearby similarities, variance is used to prioritize the ordering. 

Next, we determine the maximum value of $L$ (denoted as $L_{max}$) that ensures the performance drop remains within a given threshold $\delta$. This is achieved through a binary search and verification process, where we test the top-ranked group from $\Lambda_L$. If the verification fails, we discard the groups with more than $L$ degradations. After such coarse filtration, for each $\Lambda_L$, we further use binary search combined with training to find the group whose performance drop is closest to $\delta$, and discard all groups after it. Finally, we can obtain the feasible solution space of set $\Lambda_L$: $\Lambda_L = \{C||C|=L\}$, 
where $L=1,2,\ldots, L_{max}$. Through this preprocessing, each group in $\Lambda_L$ potentially satisfies the requirement of performance drop threshold $\delta$. This allows us to efficiently solve the partitioning scheme $\Phi$ without the need for laborious training and testing procedures.

\noindent\textbf{Degradation Grouping Search.}
In the specific process of solving, we continuously try the possible number $mid$ for $m$ through a binary search algorithm referring to network selection algorithm in TAG, while using depth-first search combined with the preprocessing results $\{\Lambda_L\}_{L=1}^{L_{max}}$ for verification. If the verification is successful, one solution is obtained, otherwise iteration continues until the solution is found. The overall process is described in Alg.~\ref{alg:degradation grouping search}.

\begin{algorithm}
\caption{Degradation Grouping Search}
\label{alg:degradation grouping search}
\begin{minipage}[t]{0.5\textwidth}
\begin{algorithmic}
\STATE \textbf{Input:} $\{d_k\}_{k=1}^n$, $\{\Lambda_L\}_{L=1}^n$, $\delta$
\STATE \textbf{Output:} \\ $\Phi(\text{d}_1,\text{d}_2,...,\text{d}_n)=\{\text{C}_1,\text{C}_2,...,\text{C}_m\}$
\STATE \textbf{Function} $Solve$$( \widetilde{C},\widetilde{\Lambda}):$
\STATE \texttt{// $\widetilde{C}$: partitioned degradations \\// $\widetilde{\Lambda}$: current groups}
\IF{$\widetilde{C}=\{d_k\}_{k=1}^n$ \textbf{and} $|\widetilde{\Lambda}|=mid$}
    \IF{$\forall C \in \widetilde{\Lambda}, \Delta P(C) \leq \delta$}
        \STATE \textbf{return} $\widetilde{\Lambda}$
    \ENDIF
\ELSE
    \FOR{$L \gets n, n-1, \ldots, 1$}
        \FOR{$C \in \widetilde{\Lambda}$}
            \IF{$|\widetilde{\Lambda}|<mid$ \textbf{and} $C \cap \widetilde{C} = \emptyset$}
                \STATE $Solve(C \cup \widetilde{C},\{C\} \cup \widetilde{\Lambda})$
            \ENDIF
        \ENDFOR
    \ENDFOR
\ENDIF
\end{algorithmic}
\end{minipage}%
\hfill
\begin{minipage}[t]{0.5\textwidth}
\begin{algorithmic}
\STATE \texttt{// Binary Search}
\STATE $left \leftarrow 1, right\leftarrow n $
\WHILE{$left < right$}
    \STATE $mid \leftarrow (left+right)/2$
    \IF{$Solve() \neq \emptyset$}
        \STATE $right \leftarrow mid$
    \ELSE
        \STATE $left \leftarrow mid+1$
    \ENDIF
\ENDWHILE
\end{algorithmic}
\end{minipage}
\end{algorithm}




Notably, the purpose of the function $Solve$ is to solve the partitioning scheme $\Phi$ when the total number of groups equals $mid$. With $\{\Lambda_L\}_{L=1}^{L_{max}}$, $Solve$ uses a depth-first search algorithm to solve all schemes that meet the constraints. Ideally, $Solve$ will not include any verification, which means that training and testing are not required. But as discussed in Sec.~\ref{sec:similarity}, the partial coupling of degradation and image content will cause some uncertainties, which makes $\{\Lambda_L\}_{L=1}^{L_{max}}$ (based on similarity matrix $S$) not completely monotonic. Therefore, we add verification in $Solve$ to ensure the correctness of the final results.

After finishing task grouping, we will obtain the partitioning schemes $\Phi^*$ which meet the requirements:

\begin{equation}
\begin{aligned}
\Phi^*(d_1,d_2,...,d_n)=\{C_1, C_2, \ldots, C_m\}.
\end{aligned}
\end{equation}

Note that there may be more than one set of solutions and the search algorithm can output all the feasible solutions. In the experiment, we only consider the first set of solution. With the grouping results, we obtain the corresponding models trained on each group. This process is concomitantly achieved by the grouping searching procedure. These grouped-training models can well deal with the degradations within its group, yielding comparable performance to the single-task upper bound models.

\subsection{Adaptive Model Selection}

Given an input image with unknown degradation $d$, GRIDS employs an adaptive model selection mechanism to intelligently determine the most appropriate mix-training model. 

This property comes as a beneficial by-product of the statistical modeling of image degradations. We first estimate the deep degradation feature distribution of the input image via self-augmentation, which crops the image into multiple small patches to capture the comprehensive degradation profile. As depicted in Sec.~\ref{sec:similarity}, these patches are used to estimate the generalized Gaussian distribution (GGD) of the input image, denoted as $GGD(d)$. For each degradation group $C_k=\{d_{k_1}, d_{k_2}, \ldots, d_{k_{n_k}}\}$ within the partitioning scheme $\Phi^*$, we define its average GGD parameters as: $GGD(C_k)=\frac{1}{n_k}\sum_{i=1}^{n_k}{GGD(d_{k_i})}$.


Then, we calculate the KL divergence between $GGD(d)$ and $GGD(C_k)$ to identify the group $C^*$ that exhibits the most similar degradation characteristics to the input image:

\begin{equation}
\begin{aligned}
C^*=\min_{C \in \Phi(\{d_k\}_{k=1}^n)}D_{kl}(GGD(C), GGD(d)).
\end{aligned}
\end{equation}

Accordingly, the most suitable mix-training model is selected as $M^*=M_M(C^*)$.
This adaptive model selection mechanism streamlines the restoration process, eliminating the need for manual model switching, as shown in Fig.~\ref{fig:model_selection_mechanism}.




\noindent\textbf{Predicting Generalization without Inference.} Another appealing advantage of our method is its ability to predict the processing performance on any input image without actual inference. After estimating the GGD parameters of the input image, we compute the distance between the input image and the selected group. A small distance $D_{kl}(GGD(C^*), GGD(d))$ indicates a high similarity between the input degradation and one of the training degradations, suggesting that the model is likely to produce satisfactory results. Conversely, a larger divergence implies a significant dissimilarity, indicating potential challenges with out-of-distribution data. This mechanism can provide insights into how well the model is expected to handle the specific degradation. It enables users to make informed decisions regarding the applicability to real-world scenarios.

\subsection{Experimental Settings}
\noindent\textbf{Degradations and Datasets.} We evaluate the proposed GRIDS on 11 degradations, including Gaussian blur, Motion blur, R-L, SR, Ringing, Defocus blur, JPEG compression, Poisson noise, Inpainting, Gaussian noise and Salt\&pepper (SP) noise. More details about the degradations are elaborated in Supp. We randomly crop $256\times 256$ image patches from DF2K dataset for training. The above 11 degradation types are added online as inputs. We use DIV2K-Valid~\cite{DIV2K} and Urban100~\cite{Urban100} datasets for testing. 

\noindent\textbf{Implementation Details.} We train single-task models as the upper bounds and a mix-training model involving all the degradations as the baseline model. The learning rate is initialized as $2e^{-4}$ and updated with warmup cosine scheme. We adopt AdamW~\cite{adamw} optimizer with $\beta_1 = 0.9$ and $\beta_2 = 0.999$. The batch size is 8. A total of 600k iterations are executed. For the Degradation Similarity Matrix, we accelerated the calculation method in SRGA by parallel computing, so that each degradation only takes 5 seconds to complete parameter estimation. For the grouping search algorithm, we employ Restormer backbone and set the performance threshold $\delta$=0.7dB, which is considered acceptable. After grouping search, the algorithm divides the 11 degradations into 4 groups: (1) $C_1$: Gaussian blur, Motion blur, R-L; (2) $C_2$: SR, Ringing, Defocus blur; (3) $C_3$: JPEG compression, Poisson noise, Inpainting; (4) $C_4$: Gaussian noise and Salt\&pepper noise. The four corresponding models are obtained as the grouping results.


\begin{table*}[t]
\centering
\resizebox{\textwidth}{!}{%
\begin{tabular}{|c|c|ccccccccccc|c|}
\hline
\textbf{Type} &
  \textbf{Methods} &
  \textbf{Gaussian Blur} &
  \textbf{Motion Blur} &
  \textbf{R-L} &
  \textbf{SR} &
  \textbf{Ringing} &
  \textbf{Defocus Blur} &
  \textbf{JPEG} &
  \textbf{Poisson Noise} &
  \textbf{Inpainting} &
  \textbf{Gaussian Noise} &
  \textbf{S\&P Noise} &
  \textbf{Avg.} \\ \hline
 &
  RealESRNet\textsuperscript{$\dagger$} &
  23.64 &
  21.39 &
  23.98 &
  25.46 &
  25.75 &
  24.70 &
  26.09 &
  26.32 &
  16.00 &
  25.20 &
  23.44 &
  23.82 \\
 &
  RealESRGAN\textsuperscript{$\dagger$} &
  21.70 &
  20.99 &
  21.38 &
  23.35 &
  23.62 &
  23.07 &
  23.07 &
  23.77 &
  24.13 &
  15.78 &
  23.26 &
  22.13 \\
 &
  AirNet\textsuperscript{$\dagger$} &
  20.68 &
  20.58 &
  22.66 &
  26.99 &
  24.30 &
  24.14 &
  38.39 &
  32.86 &
  16.27 &
  29.79 &
  17.90 &
  24.96 \\
 &
  IDR\textsuperscript{$\star$} &
  26.89 &
  25.25 &
  33.68 &
  34.52 &
  31.25 &
  36.56 &
  30.57 &
  32.77 &
  29.72 &
  29.54 &
  41.73 &
  32.04 \\
 &
  RealESRNet\textsuperscript{$\star$} &
  27.16 &
  24.88 &
  36.14 &
  33.91 &
  32.75 &
  39.73 &
  31.40 &
  33.24 &
  31.55 &
  30.04 &
  44.46 &
  33.21 \\
\multirow{-5}{*}{Mix-training} &
  AMIRNet\textsuperscript{$\star$} &
  24.54 &
  25.44 &
  30.48 &
  35.27 &
  30.89 &
  32.33 &
  31.09 &
  32.74 &
  31.12 &
  29.62 &
  39.78 &
  31.21 \\ \hline  \hline
Single-task &
  SRResNet (upper bound) &
  26.97 &
  25.36 &
  34.21 &
  34.62 &
  32.76 &
  40.15 &
  31.37 &
  33.06 &
  31.70 &
  30.08 &
  44.93 &
  33.20 \\ \cline{1-2}
Mix-training &
  SRResNet (baseline) &
  25.92 &
  22.61 &
  33.34 &
  33.70 &
  30.71 &
  37.07 &
  30.83 &
  32.51 &
  30.57 &
  29.62 &
  43.79 &
  31.88 \\ \hline
 &
  \textbf{GRIDS (SRResNet)} &
  26.71 &
  25.09 &
  34.77 &
  35.16 &
  32.54 &
  39.95 &
  31.29 &
  33.13 &
  31.90 &
  30.22 &
  45.73 &
  33.32 \\
 &
  $\Delta P$ vs. upper bound &
  \cellcolor{gray!20}-0.26 &
  \cellcolor{gray!20}-0.27 &
  \cellcolor{green!15}+0.56 &
  \cellcolor{green!15}+0.54 &
  \cellcolor{gray!20}-0.22 &
  \cellcolor{gray!20}-0.20 &
  \cellcolor{gray!20}-0.85 &
  \cellcolor{green!15}+0.07 &
  \cellcolor{green!15}+0.20 &
  \cellcolor{green!15}+0.14 &
  \cellcolor{green!15}+0.80 &
  \cellcolor{green!15}\textbf{+0.05} \\
\multirow{-3}{*}{Grouped-training} &
  $\Delta P$ vs. baseline &
  \cellcolor{green!15}+0.79 &
  \cellcolor{green!15}+2.48 &
  \cellcolor{green!15}+1.43 &
  \cellcolor{green!15}+1.46 &
  \cellcolor{green!15}+1.83 &
  \cellcolor{green!15}+2.88 &
  \cellcolor{green!15}+0.46 &
  \cellcolor{green!15}+0.62 &
  \cellcolor{green!15}+1.33 &
  \cellcolor{green!15}+0.60 &
  \cellcolor{green!15}+1.27 &
  \cellcolor{green!15}\textbf{+1.44} \\ \hline \hline
Single-task &
  Restormer (upper bound) &
  \textbf{\textcolor{blue}{27.99}} &
  \textbf{\textcolor{red}{31.80}} &
  \textbf{\textcolor{red}{37.92}} &
  \textbf{\textcolor{blue}{37.42}} &
  \textbf{\textcolor{red}{34.17}} &
  \textbf{\textcolor{red}{41.81}} &
  \textbf{\textcolor{blue}{32.14}} &
  \textbf{\textcolor{blue}{33.83}} &
  \textbf{\textcolor{blue}{32.55}} &
  \textbf{\textcolor{blue}{30.64}} &
  \textbf{\textcolor{blue}{46.12}} &
  \textbf{\textcolor{blue}{35.13}} \\ \cline{1-2}
Mix-training &
  Restormer (baseline) &
  27.39 &
  26.48 &
  35.36 &
  34.10 &
  32.29 &
  38.91 &
  31.30 &
  32.96 &
  31.56 &
  29.77 &
  42.71 &
  32.98 \\ \hline
 &
  \textbf{GRIDS (Restormer)} &
  \textbf{\textcolor{red}{28.22}} &
  \textbf{\textcolor{blue}{31.71}} &
  \textbf{\textcolor{blue}{37.24}} &
  \textbf{\textcolor{red}{38.68}} &
  \textbf{\textcolor{blue}{34.03}} &
  \textbf{\textcolor{blue}{41.30}} &
  \textbf{\textcolor{red}{32.23}} &
  \textbf{\textcolor{red}{33.88}} &
  \textbf{\textcolor{red}{32.86}} &
  \textbf{\textcolor{red}{30.71}} &
  \textbf{\textcolor{red}{46.57}} &
  \textbf{\textcolor{red}{35.22}} \\
 &
  $\Delta P$ vs. upper bound &
  \cellcolor{green!15}+0.23 &
  \cellcolor{gray!20}-0.09 &
  \cellcolor{gray!20}-0.68 &
  \cellcolor{green!15}+1.26 &
  \cellcolor{gray!20}-0.14 &
  \cellcolor{gray!20}-0.51 &
  \cellcolor{green!15}+0.09 &
  \cellcolor{green!15}+0.05 &
  \cellcolor{green!15}+0.31 &
  \cellcolor{green!15}+0.07 &
  \cellcolor{green!15}+0.45 &
  \cellcolor{green!15}\textbf{+0.09} \\
\multirow{-3}{*}{Grouped-training} &
  $\Delta P$ vs. baseline &
  \cellcolor{green!15}+0.83 &
  \cellcolor{green!15}+5.23 &
  \cellcolor{green!15}+1.88 &
  \cellcolor{green!15}+4.58 &
  \cellcolor{green!15}+1.74 &
  \cellcolor{green!15}+2.39 &
  \cellcolor{green!15}+0.93 &
  \cellcolor{green!15}+0.92 &
  \cellcolor{green!15}+1.30 &
  \cellcolor{green!15}+0.94 &
  \cellcolor{green!15}+3.86 &
  \cellcolor{green!15}\textbf{+2.24} \\ \hline
\end{tabular}
}
\caption{Quantitative comparison (PSNR) on DIV2K-Valid dataset. The proposed grouped-learning method GRIDS significantly improves the performance compared to the mix-training baseline model, while achieving comparable performance compared to the single-task upper bound models. The best and the second best performance are highlighted in \textbf{\textcolor{red}{red}} and \textbf{\textcolor{blue}{blue}} colors, respectively. \colorbox{green!15}{Green} or \colorbox{gray!20}{gray} background colors represent performance improvement or decline. {$\star$}: retrain with our dataset. {$\dagger$}: test with publicly released model.}
\label{tab:results_div2k}
\end{table*}

\section{Experiments}
\label{sec:Experiments}

\subsection{Comparison with State-of-the-art Methods}

We compare GRIDS with state-of-the-art all-in-one image restoration models, including RealESRNet, RealESRGAN~\cite{realesrgan}, AirNet~\cite{AirNet}, IDR~\cite{IDR} and AMIRNet~\cite{zhang2023all}. For RealESRGAN and AirNet, we directly use their released models for testing. RealESRNet, IDR and AMIRNet are retrained with our dataset for a fair comparison.  To verify the effectiveness and adaptability of our method, we utilize two widely-recognized network structures: SRResNet\footnote{Modified to keep the input and output resolutions the same.} and Restormer.

\begin{figure*}[htbp]
  \centering
   \includegraphics[width=\linewidth]{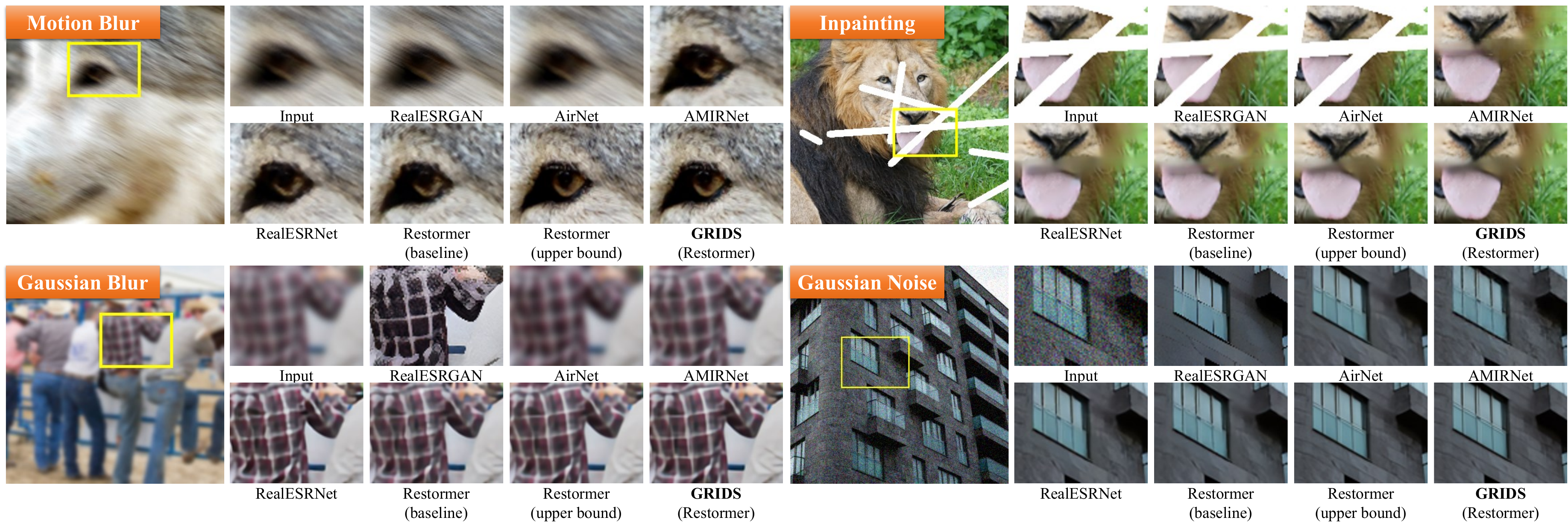}
   \caption{Visual comparisons on various degradation restoration tasks. GRIDS yields significant improvements over other methods.}
   \label{fig:visual_comparison}
\end{figure*}

From Tab.~\ref{tab:results_div2k}, it can be observed that simple mix-training methods exhibit severe performance deterioration, significantly inferior to their single-task upper bound models. For example, the mix-training SRResNet model shows an average PSNR decrease of 1.32dB and a maximum decrease of 3.08dB, while the mix-training Restormer model experiences an average PSNR decrease of 2.15dB and a maximum decrease of 5.32dB. By equipping SRResNet with GRIDS, we achieve a 1.44dB improvement over the baseline and even surpass its upper bound by 0.05dB. For the Restormer backbone, GRIDS successfully ensures that the performance gap for each task remains within the given threshold of $\delta$ = 0.7dB. It improves the baseline model by 2.24dB and outperforms its upper bound by 0.09dB. These results suggest that by appropriately grouping similar tasks, the model can learn shared representations that benefit all tasks. Visual comparisons (Fig.~\ref{fig:visual_comparison}) highlight GRIDS's proficiency in handling various degradations, producing results comparable to specialized single-task models. 
\begin{figure}[t]
      \centering
      \begin{subfigure}{0.53\textwidth}
       \includegraphics[width=\textwidth]{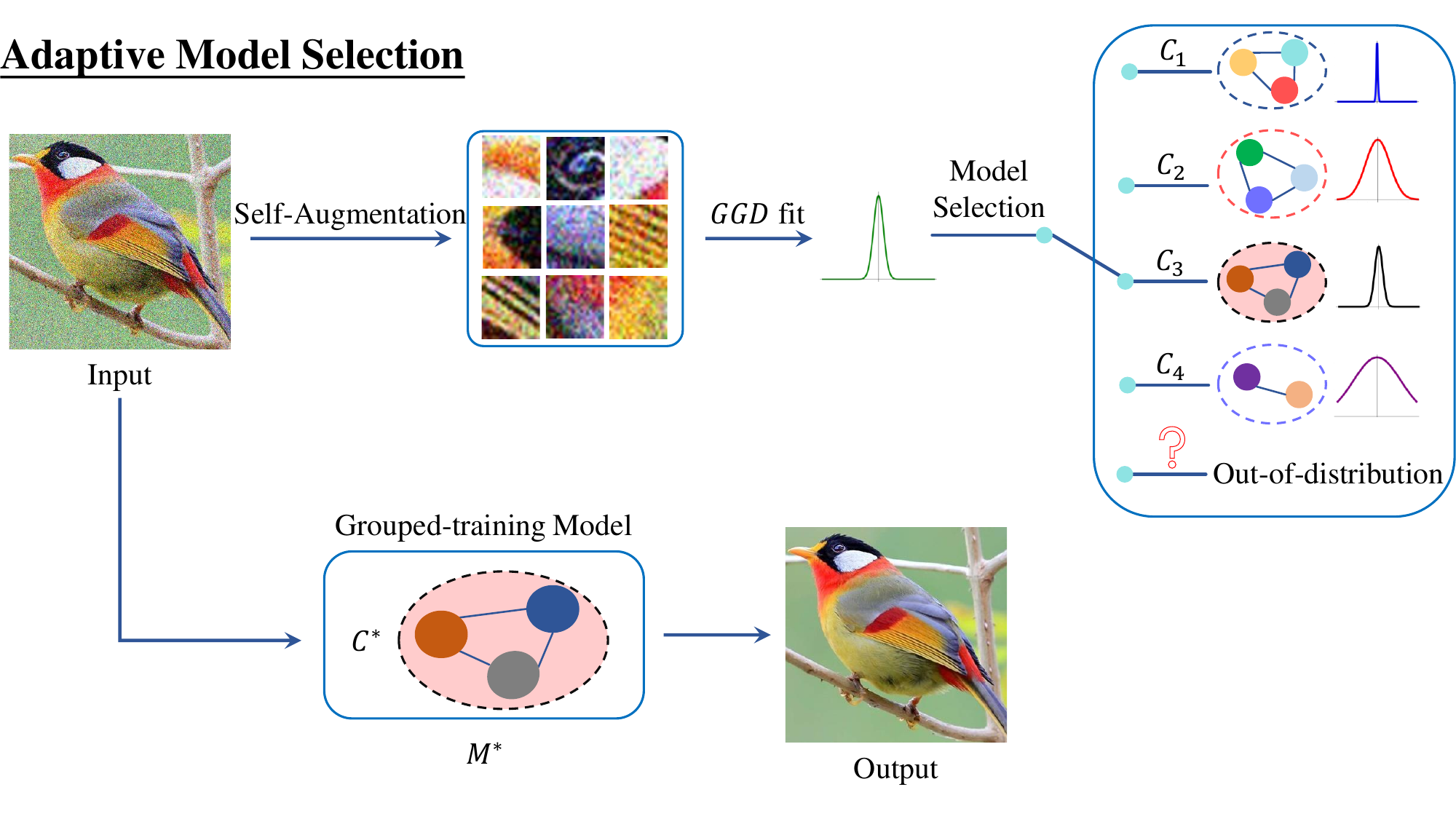}
       
       \caption{Adaptive model selection mechanism}
       \label{fig:model_selection_mechanism}
      \end{subfigure}
      \centering
      \begin{subfigure}{0.44\textwidth}
       \includegraphics[width=\textwidth]{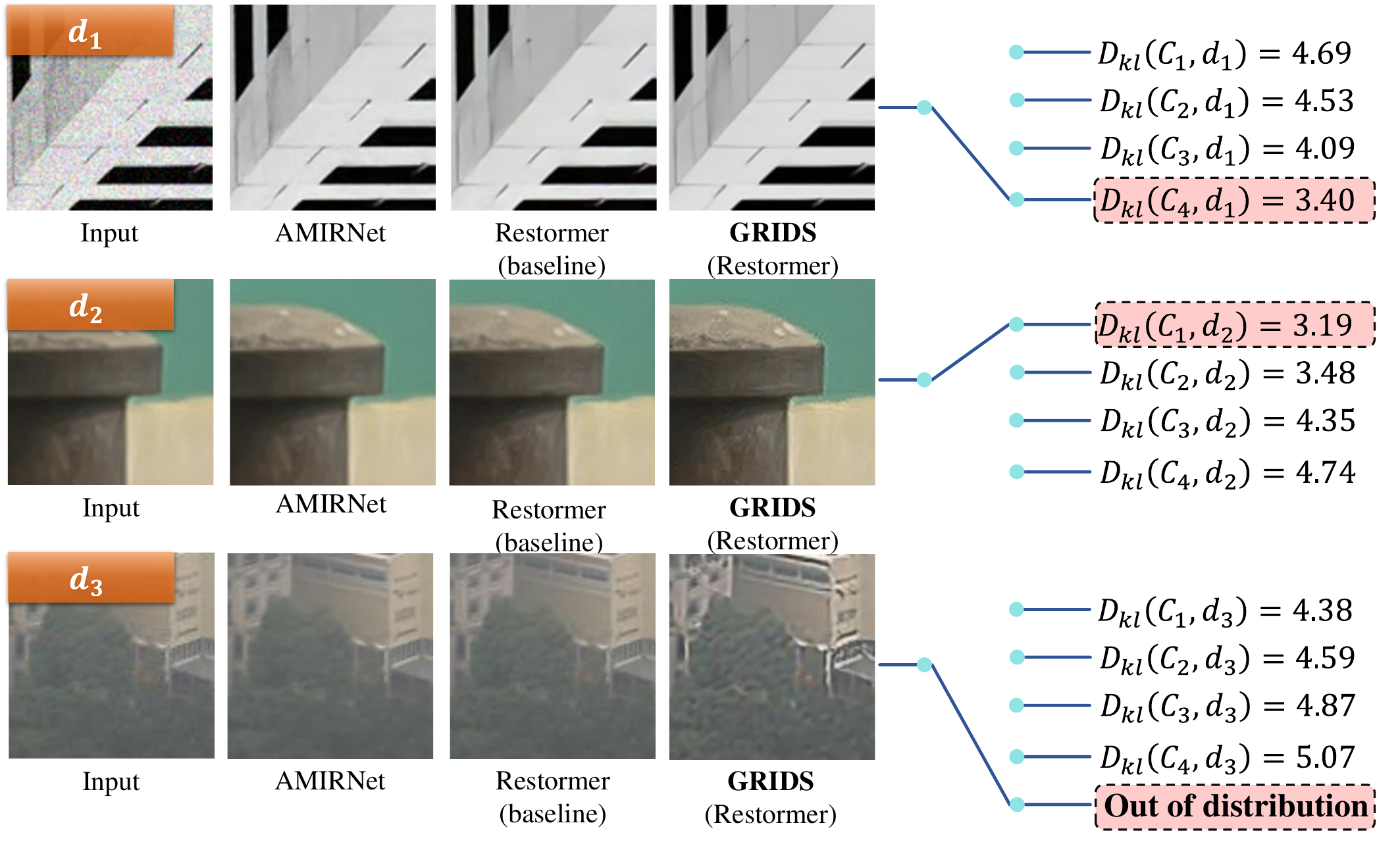}

       \caption{Examples for real-world degradations}
       \label{fig:verify_model_selection}
       \end{subfigure}
    \caption{Thanks to the statistical modeling of degradations, GRIDS can automatically switch the optimal model for an unknown input image without any auxiliary classification module. By computing the KL divergence between the input image and predefined degradation groups, it can identify the most similar group for restoration and predicts model generalization ability and processing performance without actual inference.}
    \label{fig:model_selection}
    
\end{figure}


\subsection{Effectiveness of Adaptive Model Selection}
We further validate the effectiveness of the proposed adaptive model selection mechanism. We feed the network with three different types of input images: (1) Gaussian noise images, (2) Real-world blurry images from RealSR dataset~\cite{cai2019toward}, and (3) Real-world hazy images. The experimental results are shown in Fig.~\ref{fig:verify_model_selection}. Our method successfully identifies the Gaussian noise image as belonging to group $C_4$ (Gaussian noise and Salt\&pepper noise) due to the lowest divergence with this group. The real-world blurry image is classified into group $C_1$ (Gaussian blur, Motion blur, R-L), which encompasses similar degradations associated with blurring. However, as haze degradation is not included in the training data, the divergence ($>$4) indicates that the hazy image falls outside the known distribution, rendering the current model unable to handle it. This validate the significant potential for real-world applications. More results are in the Supp.

\subsection{Effectiveness of Degradation Similarity and Limitations}
Based on the degradation similarity matrix obtained in Sec.~\ref{sec:similarity}, we devise an automatic searching algorithm to divide the various degradations into related groups. In this section, we further validate the effectiveness of the degradation similarity. By utilizing the average similarity, we designed multiple sets of experiments, each of them includes a group of degradations with higher similarity (indicated by lower distance) and another group with lower similarity for mixed training. The experiments cover different degradation numbers and similarity gaps, provided in Tab.~\ref{tab:similarity_DDR_effective}. The results reveal that GRIDS performs well on these cases. For example, the group in Experiment 2 with higher similarity (dist: 2.20) exhibits a smaller performance drop, while the group with lower similarity (dist: 4.33) experiences a more substantial performance drop (up to a 1.72dB decline for S\&P Noise). More comprehensive analysis is in the Supp.

\noindent\textbf{Comparison with other alternative features.} Besides the DDR features, there may be other alternatives to derive the degradation similarity, such as using deep VGG~\cite{vgg} features or artificially designed SIFT~\cite{sift} features. We modify only the features used in estimating GGD parameters, while keeping the rest of the calculation procedure the same. The experimental results, presented in Tab. ~\ref{tab:similarity_VGG_SIFT}, indicate that using VGG and SIFT features to estimate the degradation similarity is not as accurate. The performance drop of high-similarity degradations is larger compared to that of low-similarity degradations. For example, the performance drop of SR task surpasses 2.5dB in the higher-similarity group, which violates the relationships identified by VGG and SIFT. This discrepancy underscores the superiority and precision of our proposed DDR-based similarity measure. Further analysis is in the Supp.

\noindent\textbf{Limitations.} DDR-based Degradation Similarity can accurately reflect the proximity of degradation in most cases, but errors may occur for groups with a small similarity gap. As shown in Tab.~\ref{tab:similarity_DDR_limitation}, for two groups with a similarity gap of 0.12, the average performance drop may surpass 0.1dB in the higher-similarity group. However, the aforementioned errors exert minimal influence and have been duly considered within our algorithm, thereby not impacting the final results.

\begin{table}[t]
\centering
\resizebox{\textwidth}{!}{%
\begin{tabular}{cc}
\begin{minipage}{0.45\textwidth}
    \centering
    \begin{subtable}[t]{\linewidth}
        \centering
        \captionsetup{labelformat=parens, labelsep=quad, skip=5pt}
        \resizebox{\linewidth}{!}{
                \begin{tabular}{|c|c|c|c|} 
                \hline
                \multicolumn{4}{|c|}{\textbf{Experiment 1:} 2 degradations, similarity gap=0.79} \\ \hline
                \textbf{Higher similarity (dist: 2.75)} & \textbf{SR}       &  \textbf{Defocus Blur} & \textbf{Avg.}           \\ \hline
                upper bound              &  37.42         & 41.81       &  39.62        \\ \hline
                mix-training             &  39.22         & 41.38       &  40.30         \\ \hline
                \cellcolor{green!15}$\Delta P$               &  \cellcolor{green!15}1.80         & \cellcolor{green!15}-0.43       & \cellcolor{green!15}\textbf{\textcolor{red}{0.68}} \\ \hline 
                \textbf{Lower similarity (dist: 3.54)}  & \textbf{Gaussian Noise} & \textbf{S\&P Noise}  & \textbf{Avg.}           \\ \hline
                upper bound              &  30.64        & 46.12       & 38.38          \\ \hline
                mix-training             &  30.71        & 46.57       & 38.64         \\ \hline
                \cellcolor{green!15}$\Delta P$               &  \cellcolor{green!15}0.07        & \cellcolor{green!15}0.45       & \cellcolor{green!15}\textbf{\textcolor{blue}{0.26}} \\ \hline
                \end{tabular}
            }
            \resizebox{\linewidth}{!}{
                \begin{tabular}{|c|c|c|c|c|}
                \hline
                \multicolumn{5}{|c|}{\textbf{Experiment 2:} 3 degradations, similarity gap=2.13} \\ \hline
                \textbf{Higher similarity (dist: 2.20)} & \textbf{Ringing}       & \textbf{R-L}   & \textbf{Defocus Blur} & \textbf{Avg.}           \\ \hline
                upper bound              & 34.17         & 37.92 & 41.81       & 37.97          \\ \hline
                mix-training             & 34.10         & 37.66 & 41.25       & 37.67          \\ \hline
                \cellcolor{green!15}$\Delta P$               & \cellcolor{green!15}-0.08         & \cellcolor{green!15}-0.26 & \cellcolor{green!15}-0.56       & \cellcolor{green!15}\textbf{\textcolor{red}{-0.30}} \\ \hline 
                \textbf{Lower similarity (dist: 4.33)}  & \textbf{Gaussian Blur} & \textbf{JPEG}  & \textbf{S\&P Noise}  & \textbf{Avg.}           \\ \hline
                upper bound              & 27.99         & 32.14 & 46.12       & 35.42          \\ \hline
                mix-training             & 27.95         & 31.77 & 44.40       & 34.71          \\ \hline
                \cellcolor{green!15}$\Delta P$               & \cellcolor{green!15}-0.04         & \cellcolor{green!15}-0.37 & \cellcolor{green!15}-1.72       & \cellcolor{green!15}\textbf{\textcolor{blue}{-0.71}} \\ \hline
                \end{tabular}
            }
            \resizebox{\linewidth}{!}{
                \begin{tabular}{|c|c|c|c|c|c|}
                \hline
                \multicolumn{6}{|c|}{\textbf{Experiment 3:} 4 degradations, similarity gap=1.58} \\ \hline
                \textbf{Higher similarity (dist: 2.52)}   & \textbf{Ringing}   & \textbf{R-L} & \textbf{Defocus Blur} & \textbf{Motion Blur} & \textbf{Avg.}           \\ \hline
                upper bound              &  34.17        & 37.92 &  41.81     & 31.80  & 36.43       \\ \hline
                mix-training             &  33.37        & 36.74 & 40.33  & 29.66 &  35.03    \\ \hline
                \cellcolor{green!15}$\Delta P$               &  \cellcolor{green!15}-0.80        & \cellcolor{green!15}-1.18      & \cellcolor{green!15}-1.48  &  \cellcolor{green!15}-2.14 & \cellcolor{green!15}\textbf{\textcolor{red}{-1.40}} \\ \hline
                \textbf{Lower similarity (dist: 4.10)}  & \textbf{Gaussian Blur} & \textbf{Gaussian Noise}  & \textbf{S\&P Noise} & \textbf{Motion Blur} & \textbf{Avg.}           \\ \hline
                upper bound              &  27.99        & 30.64 & 46.12      & 31.80   &  34.14    \\ \hline
                mix-training             &   27.89       & 30.28       & 44.20  & 26.66  & 32.26     \\ \hline
                \cellcolor{green!15}$\Delta P$               &  \cellcolor{green!15}-0.10        & \cellcolor{green!15}-0.36       & \cellcolor{green!15}-1.92  &  \cellcolor{green!15}-5.14 & \cellcolor{green!15}\textbf{\textcolor{blue}{-1.88}} \\ \hline 
                \end{tabular}
            }
            \resizebox{\linewidth}{!}{
                \begin{tabular}{|c|c|c|c|c|c|c|}
                \hline
                \multicolumn{7}{|c|}{\textbf{Experiment 4:} 5 degradations, similarity gap=1.12} \\ \hline
                \textbf{Higher similarity (dist: 2.88)} & \textbf{SR}  & \textbf{JPEG}     & \textbf{Ringing} & \textbf{R-L}   & \textbf{Defocus Blur} & \textbf{Avg.}           \\ \hline
                upper bound              &  37.42         & 32.14      & 34.17 & 37.92  & 41.81 & 36.70     \\ \hline
                mix-training             &  38.55       & 31.56      & 33.96  & 36.51 &  41.18 &  36.35  \\ \hline
                \cellcolor{green!15}$\Delta P$               &  \cellcolor{green!15}1.13         & \cellcolor{green!15}-0.58      & \cellcolor{green!15}-0.21  & \cellcolor{green!15}-1.41 & \cellcolor{green!15}-0.63 & \cellcolor{green!15}\textbf{\textcolor{red}{-0.34}} \\ \hline 
                \textbf{Lower similarity (dist: 4.00)} & \textbf{Gaussian Blur}   & \textbf{Inpainting} & \textbf{S\&P Noise}  & \textbf{Poisson Noise} & \textbf{Motion Blur} & \textbf{Avg.}           \\ \hline
                upper bound              &  27.99 &  32.55        & 46.12      & 33.83  & 31.80 & 34.46     \\ \hline
                mix-training             &  27.77        & 32.23      & 45.20  & 33.47  & 29.22 & 33.58    \\ \hline
                \cellcolor{green!15}$\Delta P$               &  \cellcolor{green!15}-0.22        & \cellcolor{green!15}-0.32      & \cellcolor{green!15}-0.92  & \cellcolor{green!15}-0.36 & \cellcolor{green!15}-2.58 & \cellcolor{green!15}\textbf{\textcolor{blue}{-0.88}} \\ \hline 
                \end{tabular}
            }
        \caption{Experiments on different similarity gaps and degradation numbers.}
        \label{tab:similarity_DDR_effective}
        \end{subtable}
\end{minipage}
& 
\begin{minipage}{0.45\textwidth}
    \centering
    \begin{subtable}[t]{\linewidth}
        \centering
        \captionsetup{labelformat=parens, labelsep=quad, skip=5pt}
        \resizebox{\linewidth}{!}{
            \begin{tabular}{|c|c|c|c|c|}
                \hline
                \multicolumn{5}{|c|}{\textbf{Experiment 5: }VGG features} \\ \hline
                \textbf{Higher similarity (dist: 1.79)} & \textbf{SR} & \textbf{JPEG} & \textbf{Possion Noise} & \textbf{Avg.} \\ \hline
                upper bound & 37.42 & 32.14 & 33.83 & 34.46 \\ \hline
                mix-training & 34.89 & 31.82 & 33.50 & 33.40 \\ \hline
                \cellcolor{gray!20}$\Delta P$ & \cellcolor{gray!20}-2.53 & \cellcolor{gray!20}-0.32 & \cellcolor{gray!20}-0.33 & \cellcolor{gray!20}\textbf{\textcolor{blue}{-1.06}} \\ \hline 
                \textbf{Lower similarity (dist: 3.43)} & \textbf{Gaussian Blur} & \textbf{Inpainting} & \textbf{Possion Noise} & \textbf{Avg.} \\ \hline
                upper bound & 27.99 & 32.55 & 33.83 & 31.46 \\ \hline
                mix-training & 27.64 & 31.96 & 33.32 & 30.97 \\ \hline
                \cellcolor{gray!20}$\Delta P$ & \cellcolor{gray!20}-0.35 & \cellcolor{gray!20}-0.59 & \cellcolor{gray!20}-0.51 & \cellcolor{gray!20}\textbf{\textcolor{red}{-0.49}} \\ \hline
            \end{tabular}
        }
        \resizebox{\linewidth}{!}{
            \begin{tabular}{|c|c|c|c|c|}
                \hline
                \multicolumn{5}{|c|}{\textbf{Experiment 6: }SIFT features} \\ \hline
                \textbf{Higher similarity (dist: 2.11)} & \textbf{SR} & \textbf{JPEG} & \textbf{Inpainting} & \textbf{Avg.} \\ \hline
                upper bound & 37.42 & 32.14 & 32.55 & 34.03 \\ \hline
                mix-training & 34.84 & 31.37 & 31.84 & 32.68 \\ \hline
                \cellcolor{gray!20}$\Delta P$ & \cellcolor{gray!20}-2.58 & \cellcolor{gray!20}-0.77 & \cellcolor{gray!20}-0.71 & \cellcolor{gray!20}\textbf{\textcolor{blue}{-1.35}} \\ \hline
                \textbf{Lower similarity (dist: 4.42)} & \textbf{Gaussian Blur} & \textbf{Ringing} & \textbf{Defocus Blur} & \textbf{Avg.} \\ \hline
                upper bound & 27.99 & 34.17 & 41.81 & 34.66 \\ \hline
                mix-training & 27.80 & 33.94 & 41.15 & 34.30 \\ \hline
                \cellcolor{gray!20}$\Delta P$ & \cellcolor{gray!20}-0.19 & \cellcolor{gray!20}-0.23 & \cellcolor{gray!20}-0.66 & \cellcolor{gray!20}\textbf{\textcolor{red}{-0.36}} \\ \hline
            \end{tabular}
        }
        \caption{VGG and SIFT features cannot accurately reflect the degradation relationships.}
        \label{tab:similarity_VGG_SIFT}
    \end{subtable}
    
    \centering
    \begin{subtable}[t]{\linewidth}
        \centering
        \captionsetup{labelformat=parens, labelsep=quad, skip=5pt}
        \resizebox{\linewidth}{!}{
                \begin{tabular}{|c|c|c|c|c|}
                \hline
                \multicolumn{5}{|c|}{\textbf{Experiment 7:} 3 degradations, similarity gap=0.12} \\ \hline
                \textbf{Higher similarity (dist: 3.49)} & \textbf{JPEG}  & \textbf{Poisson Noise}      & \textbf{Gaussian Noise} & \textbf{Avg.}           \\ \hline
                upper bound              &  32.14                & 33.83  & 30.64 &  32.20   \\ \hline
                mix-training             &  32.24              & 33.88  & 30.65 & 32.26     \\ \hline
                \cellcolor{gray!20}$\Delta P$               &  \cellcolor{gray!20}0.10               & \cellcolor{gray!20}0.05  & \cellcolor{gray!20}0.01 & \cellcolor{gray!20}\textbf{\textcolor{blue}{0.05}} \\ \hline 
                \textbf{Lower similarity (dist: 3.61)}  & \textbf{JPEG}  & \textbf{Poisson Noise}      & \textbf{Inpainting} & \textbf{Avg.}           \\ \hline
                upper bound              &  32.14              & 33.83  & 32.55 &  32.84    \\ \hline
                mix-training             &  32.23              & 32.86  & 33.88 & 32.99     \\ \hline
                \cellcolor{gray!20}$\Delta P$               & \cellcolor{gray!20}0.09               & \cellcolor{gray!20}-0.97 & \cellcolor{gray!20}1.33  & \cellcolor{gray!20}\textbf{\textcolor{red}{0.15}} \\ \hline 
                \end{tabular}
            }
        \caption{Failure cases may occur under close similarity scores.}
        \label{tab:similarity_DDR_limitation}
    \end{subtable}
\end{minipage}
\\
\end{tabular}
}\
\caption{(a) \& (b) Our method can quantitatively measure the relationships between degradations, while VGG and SIFT features fails to characterize the degradation representations. (c) When the distance between degradations is relatively close, our method may experience fluctuations, yet GRIDS is still capable of finding solutions that meet the requirements. The higher and lower average performance drop are highlighted in \textbf{\textcolor{red}{red}} and \textbf{\textcolor{blue}{blue}} colors, respectively. \colorbox{green!15}{Green} or \colorbox{gray!20}{gray} background colors represent whether the feature accurately reflects the degradation similarities.}
\label{tab:effectiveness_limitations_DDR}
\end{table}

\section{Conclusion}
\label{sec:Conclusion}
In this paper, we provide a new method for multiple-degradation image restoration called GRIDS. We quantitatively describe the degradation similarity from a statistical perspective, and partition the degradations into the most efficient groups through a degradation grouping search. We perform mix-training based on the partitioning results to obtain the final models, which show excellent performance on up to 11 different degradation types. Compared with other models, GRIDS can greatly improve the performance in multiple-degradation restoration. In addition, GRIDS has the unique ability to adaptively select the optimal grouped-training model and provide generalization estimation in advance.


\clearpage  

\section*{Acknowledgements}
This work was supported by Shanghai Artificial Intelligence Laboratory and 
National Natural Science Foundation of China (Grant No.62276251, 62272450), and the Joint Lab of CAS-HK.

%
%
\bibliographystyle{splncs04}
\bibliography{main}
\end{document}